%% file: main.tex
\documentclass[conference]{IEEEtran}
\usepackage[utf8]{inputenc}
\IEEEoverridecommandlockouts
\usepackage{amsmath,amssymb,amsfonts}
\usepackage{algorithmic}
\usepackage{graphicx}
\usepackage{subfig}
\usepackage{rotating}
\usepackage{textcomp}
\usepackage{xcolor}
\usepackage{hyperref}
\usepackage[inline]{enumitem}
\usepackage{amssymb}

\def\BibTeX{{\rm B\kern-.05em{\sc i\kern-.025em b}\kern-.08em
    T\kern-.1667em\lower.7ex\hbox{E}\kern-.125emX}}

\input{pre_text_cmd.tex}    
\input{pre_code.tex}
\input{pre_bib.tex}

\let\svthefootnote\thefootnote
\newcommand\blankfootnote[1]{%
  \let\thefootnote\relax\footnotetext{#1}%
  \let\thefootnote\svthefootnote%
}

\begin{document}

\title{Towards a computer-interpretable actionable formal model to encode data governance rules
\thanks{This work has been accepted and should appear in the Proceedings of IEEE eScience 2019 Conference (BC2DC). Please cite the published work instead of this one when possible.}
}

\author{\IEEEauthorblockN{Rui Zhao}
\IEEEauthorblockA{\textit{School of Informatics} \\
\textit{University of Edinburgh}\\
Edinburgh, UK \\
s1623641@sms.ed.ac.uk}
\and
\IEEEauthorblockN{Malcolm Atkinson}
\IEEEauthorblockA{\textit{School of Informatics} \\
\textit{University of Edinburgh}\\
Edinburgh, UK \\
Malcolm.Atkinson@ed.ac.uk}
}

\maketitle

\begin{abstract}
With the needs of science and business, data sharing and re-use has become an intensive activity for various areas. In many cases, governance imposes rules concerning data use, but there is no existing computational technique to help data-users comply with such rules. We argue that intelligent systems can be used to improve the situation, by recording provenance records during processing, encoding the rules and performing reasoning. We present our initial work, designing formal models for \tterm{data rules} and \tterm{flow rules} and the reasoning system, as the first step towards helping data providers and data users sustain productive relationships.
\end{abstract}


\input{sec_introduction.tex}

\input{sec_related_research.tex}

\input{sec_lang.tex}

\input{sec_system.tex}

\input{sec_contribution.tex}

\input{sec_conclusion.tex}

\printbibliography

\end{document}

%% file: pre_text_cmd.tex
\newcommand{\res}{\emph}

\newcommand{\tkw}{\textbf}

\newcommand{\tterm}{\emph}

\newcommand{\trule}{\emph}

%% file: pre_code.tex
\usepackage{listings}
\usepackage{algorithm,algorithmic}

\lstdefinestyle{pseudo}{
  belowcaptionskip=1\baselineskip,
  breaklines=true,
  frame=L,
  xleftmargin=\parindent,
  language=C,
  showstringspaces=false,
  basicstyle=\footnotesize\ttfamily,
  keywordstyle=\bfseries\color{green!40!black},
  commentstyle=\itshape\color{purple!40!black},
  identifierstyle=\color{blue},
  stringstyle=\color{orange},
}

\lstdefinestyle{datarule}{
  belowcaptionskip=1\baselineskip,
  breaklines=true,
  frame=L,
  xleftmargin=\parindent,
  showstringspaces=false,
  basicstyle=\footnotesize\ttfamily,
}

\newcommand{\ttdrin}{\lstinline[style=datarule]}

\newcommand{\LET}[2]{\STATE #1 $\gets$ #2}

%% file: pre_bib.tex
\usepackage[english]{babel}
\usepackage{csquotes}
\usepackage[style=ieee,maxnames=5,isbn=false]{biblatex}

\DeclareSourcemap{
  \maps[datatype=bibtex, overwrite=true]{
    \map{
      \step[fieldset=keywords, fieldvalue=offline]
    }
    \map{
      \pertype{online}
      \step[fieldset=keywords, fieldvalue=online]
    }
    \map{
      \pertype{misc}
      \step[fieldsource=url, final]
      \step[fieldset=keywords, fieldvalue=online]
    }
    \map{
      \step[fieldsource=keywords, match=offline, final]
      \step[fieldset=url, null]
      \step[fieldset=urldate, null]
    }
  }
}

%% file: sec_introduction.tex
\section{Introduction}
\label{sec:intro}

Data ethics and privacy are of rising importance, especially with the establishment of GDPR \cite{ujcich_provenance_2018}. Similar issues also apply in research when data from various sources are used as inputs to analyses and simulations. Researchers are aware that there are governance rules applied to the data, but they can easily lose track of the rules when the number of sources becomes large. The large volume of rules brings problem from three aspects:

\begin{enumerate}
    \item to fully read and understand the rules;
    \item to consider the consequence of combining data and their associate rules;
    \item to assign rules to output so that results can be used compliantly.
\end{enumerate}

\noindent One response is to make data open and freely accessible (e.g.~under the FAIR principle \cite{wilkinson_fair_2016} and/or as Linked Open Data as suggested by Tim Berners-Lee \cite{tim_berners-lee_linked_2009}). This sounds nice but it still leaves rules, for example to properly acknowledge sources and to protect personal and commercially sensitive data, even within collaborating communities \cite{demchenko_cyclone:_2016}. Moreover, this doesn't solve (or even decrease) the prevalent \emph{polarization}: data are either completely public (with one or a few well-known commonly agreed governance rules) or completely under control with heterogeneous (yet potentially similar) governance rules written in different languages, similar to the situation for copyright licenses.

This issue becomes more serious when IoT devices (especially sensors) are widely used: data from them can be more sensitive, but users have limited control over where the data will go \cite{sicari_security_2015}\cite{islam_internet_2015}. 
Therefore, it is necessary to let intelligent systems handle this as much as possible, while still getting people involved and ensuring they understand what is needed. % as much as possible.

We propose to pioneer a combination of technology and its modes of use to help providers and users of data communicate precisely about the rules. We also set out to enable computer systems aid in compliance with rules while processing data. This would use automation to draw attention to rules at the relevant moments and collect information to support compliance with the rules. This requires that the notations are sufficiently machine readable and detailed to support the automation, and can be directly or indirectly (through automatic conversion) authored, understood and used by humans.

There are three approaches possible:

\begin{enumerate}
    \item to police the system excessively taking little account of semantic details in order to prohibit every possible violation. This often excludes valid activities.
    \item interpret the users' actions and inputs to verify in detail that rules will be honored. This is beyond the state of the art when users have the full power of the system.
    \item encode the rules taking the semantics into account, and annotate the processing where necessary to perform reasoning. This requires extra work for both the data providers and the process developers.
\end{enumerate}

\noindent Our work focuses on the third category which requires solutions to three issues:

\begin{enumerate}
    \item \label{enum:tgt0:lang} How to write the governance rules in a computer-interpretable language?
    \item \label{enum:tgt0:check} How to handle the compliance checking?
    \item \label{enum:tgt0:verify} How to ensure that malevolent actors cannot circumvent the rules?
\end{enumerate}

\noindent Our work deals with the first two aspects, while the third is future work as it involves security and protection (for which research like \cite{missier_mind_2017} may be applied). 
In addition, the fact that the data processing is done by different bodies makes the problem inherently \emph{federated}, so the solution should also be federated. Therefore, the solution needs to be:

\begin{enumerate}
    \item \label{enum:req:va} vendor- and technology-agnostic;
    \item \label{enum:req:ex} able to extend easily to different disciplines;
    \item \label{enum:req:int} interoperable across institutional and even national boundaries; and
    \item \label{enum:req:decentral} not rely on a single trust authority.
\end{enumerate}

\noindent These requirements direct our research to associate rules with data, rather than use a central system to control data access. 
A method to describe the effect of the processing steps on rules is also needed, in order to decide which rules need to be propagated or revised for the outputs. 
Therefore the solution includes the following:

\begin{enumerate}
    \item \label{enum:tgt:proc} \tkw{encoding procedure} \quad a procedure to transform natural language governance rules to a formal representation;
    \item \label{enum:tgt:lang} \tkw{rule language} \quad a computer-interpretable, extensible and interoperable formal language to write the data-governance rules;
    \item \label{enum:tgt:proto} \tkw{rule association} \quad a standard protocol to associate encoded governance rules with the data;
    \item \label{enum:tgt:dflow} \tkw{data flow representation} \quad a vendor- and technology-agnostic method to represent and record data flow;
    \item \label{enum:tgt:change} \tkw{flow behavior} \quad a mechanism to allow processes to change the propagated governance rules;
    \item \label{enum:tgt:rsn} \tkw{reasoning system} \quad a reasoning system capable of working with any data-flow topology;
    \item \label{enum:tgt:verify} \tkw{correctness verification} \quad a mechanism (or several mechanisms) to ensure or verify that all the steps above are conducted correctly.
\end{enumerate}

\noindent Specifically, targets \ref{enum:tgt:lang}, \ref{enum:tgt:change} and \ref{enum:tgt:rsn} are the main aspects that our work and standardized provenance is used as the data-flow representation.

% the series of research questions realted to this direction should be identified clearly. We identify that a series of technologies is needed to:

% \begin{enumerate}
%     \item \label{enum:tgt0:lang} write the governance rules in a computer-interpretable language;
%     \item \label{enum:tgt0:check} handle the compliance check of data processing;
%     \item \label{enum:tgt0:out} reason about the rules for the output data;
%     \item \label{enum:tgt0:verify} ensure all the steps above are conducted correctly.
% \end{enumerate}

% \noindent By this means, the data originator can specify their data governance rules, and everyone can proceed with more faith not to break the rules. Moreover, the fact that the data processing is done by different bodies makes this problem inherently \emph{federated}, which is an aspect not correctly handled for many existing research (see section \ref{sec:question}).

% In this paper, we introduce the questions in-depth, and describe the targeting aspects needed to solve the problem. We then link the related research to these targets, and then present our work towards solving the problem targeting a specific context: helping researchers to manage the rule compliance.

%In this paper, we describe our effort towards solving \ref{enum:tgt:lang}, \ref{enum:tgt:check}, \ref{enum:tgt:out}, \ref{enum:tgt:va}, \ref{enum:tgt:ex} and \ref{enum:tgt:int}: a formal model to encode data (governance) rules, a formal way of representing inner-flow and a reasoning system working on top of provenance.

The structure of this paper is as follows: section \ref{sec:related-research} introduces the related research; section \ref{sec:lang} describes the rule language we propose and provides some examples; section \ref{sec:arch} describes the design and architecture of our system, and presents initial results; section \ref{sec:contrib} highlights the contribution of our work and introduces the future work. A conclusion is drawn in section \ref{sec:concl}.

%% file: sec_related_research.tex
\section{Related research}
\label{sec:related-research}

Here we describe the related research, stating its relation with the different aspects of our targets, as well as the limitations of each approach. 
It is worth noting that much research uses the word ``policy'' instead of ``rule'', and the part of the rules they capture lies mainly in controlling access.

A (now deprecated) standard called Platform for Privacy Preferences (P3P) was established by W3C in 2002, aimed at allowing websites to specify their privacy policies and compare those against user's preference to grant or prevent access (from the user's agent, e.g. web browsers). However, P3P had a very limited vocabulary and was not widely implemented in web browsers. It was finally deprecated and no successive standards were established.

Building on some concepts from P3P, E-P3P (Platform for Enterprise Privacy Practices) \cite{karjoth_platform_2002}, one of the earliest works we found. 
It attempted to address the expressivity issue in a ``formal'' way by describing the privacy policy by specifying permitted data users, purposes and operations, and specifying consequent obligations. 
However, the set of predicates was very limited and the work payed no attention to the federated context, especially the interoperability issue. 
Checking compliance with policies associated with data was not completed there. It was reported in \cite{mont_towards_2003}; often cited as the foundation of the \emph{sticky policy}.

\res{Sticky policy} \cite{mont_towards_2003}\cite{pearson_sticky_2011} provided a conceptual framework to associate policies with data and ``ensure'' the data handlers (people, institutions, etc) are aware of the policies, by encrypting the data first, sending the policies with the encrypted data and sending the decryption key after checking that they agreed to comply with the rules. 
However, the checking procedure is not automatic and relies on the so-called Trusted Authority (some agencies explicitly trusted by the data owner). 
That said, in principle, the encrypt-and-decrypt concept can be used as the foundation of the protocol for an automated checking mechanism on each occasion and at each site he data is used.

\res{CamFlow} \cite{pasquier_camflow:_2017} is the work most similar to ours, though there is still much difference. It uses (Decentralized) Information Flow Control (IFC) \cite{myers_decentralized_1997} to represent the rules in the form of tags (in two groups, \emph{secrecy} and \emph{integrity}), which is a very simplified notation: compatibility of tags of the data and the process are checked before processing. On the other hand, CamFlow captures the importance of associating rules with data and allowing processes to change the associated rules (by specifying the modification to the output tags for each process), though it did not show any work in handling multi-input-multi-output processes. Moreover, CamFlow provides a mechanism to handle the data flowing between machines, making it possible to be used in a decentralized context.

\res{Meta-code} \cite{havard_d._johansen_enforcing_2015} is similar to CamFlow, and this technology is used in a series of works (\cites{h._d._johansen_management_2014,hurley_self-managing_2014,a._t._gjerdrum_implementing_2016}). They also use the IFC concept to assign permission tags (semantically, roles) to data, and then specify the flow behaviour (by specifying the \emph{policy file of output data}) on the governance rule side (instead of the process specification side, like in CamFlow). In addition, they have a special \emph{meta-code} part which is basically a program (resulting in either pass or error) executed at any specific file processing action (e.g. \texttt{onAccess}). However, because this is an arbitrary program, the policy of the resulting data can only be known at runtime, making it hard to do static analysis.

\res{Thoth} \cite{elnikety_thoth:_2016} on the other hand, favors the rule specification purely from the data originator side, because of its context: local data usage (\emph{search engine} is used as an example). They have a set of logical connectives and predicates to specify the read, update and declassify policies. They do not use role tags like CamFlow and Meta-code do, but they allow the use of cryptographic keys (e.g. \texttt{sKeyIs(x)}) or IP addresses (e.g. \texttt{sIpIs(x)}) to identify data users and allow a special loop structure to test the match (existence). A further work, SHAI (\cite{eslam_elnikety_shai:_2018}), provides a method to do static analysis for part of the rule language of Thoth. Despite the expressiviness, the fact that everything is based on the rules assigned by the data originators either requires them to be ``omniscient'' or prohibits many data uses that should be valid in principal.

\res{Legal modelling} is a whole different field, where precisely modelling the document is the priority. \cite{robaldo_combining_2015} and \cite{robaldo_reified_2017} uses reified predicate logic with Input/Output logic to model the legal corpus into their logical representation (e.g.~GDPR%
\footnote{dapreco, LegalRuleML formulae of deontic rules in the DAPRECO project.: Dapreco/daprecokb https://github.com/dapreco}%
). \cite{ujcich_provenance_2018} and \cite{pandit_queryable_2018} both looked at adding additional information to provenance to check the compliance of execution processes (against GDPR). However, these all focused purely on the modelling of terms, and never considered that the data will change (and so the rules will change) during processing.

There are also research and standards aimed at providing languages to express policies, with no binding to any processing systems, such as MPEG-21\footnote{ISO/IEC TR 21000-1:2004 https://www.iso.org/standard/40611.html}, XACML\footnote{eXtensible Access Control Markup Language (XACML) Version 3.0 https://docs.oasis-open.org/xacml/3.0/xacml-3.0-core-spec-os-en.html} and ODRL\footnote{ODRL Information Model 2.2 https://www.w3.org/TR/odrl-model/\#policy-agreement}.
However, every one of them aims at expressing the policy for ``the (specific) data'', paying very limited attention (if any) to the policy for ``derived data''. From our perspective, Open Digital Rights Language (ODRL) is the one most closely related to our work. It provides an extensible semantic way to describe permissions, prohibitions and obligations as well as more fine-grained constraints. But because of its design priorities, data are explicitly specified and no mechanism is provided to support policy change, which means it does not directly meet our requirements. On the other hand, since ODRL provides its vocabulary in which many concepts are defined almost the same as ours, we will reuse some of its definitions in our language in the future.

%% file: sec_lang.tex
\section{Rule language}
\label{sec:lang}

Having compared the related research and described the general context, the requirements of the governance rule language are set as follows:

\begin{enumerate}
    \item be precise enough (i.e.~have no ambiguity);
    \item be able to talk about more than just access controls;
    \item be interoperable across institutional (and jurisdictional) boundaries.
\end{enumerate}

\begin{figure}
    \centering
    \includegraphics[width=\linewidth]{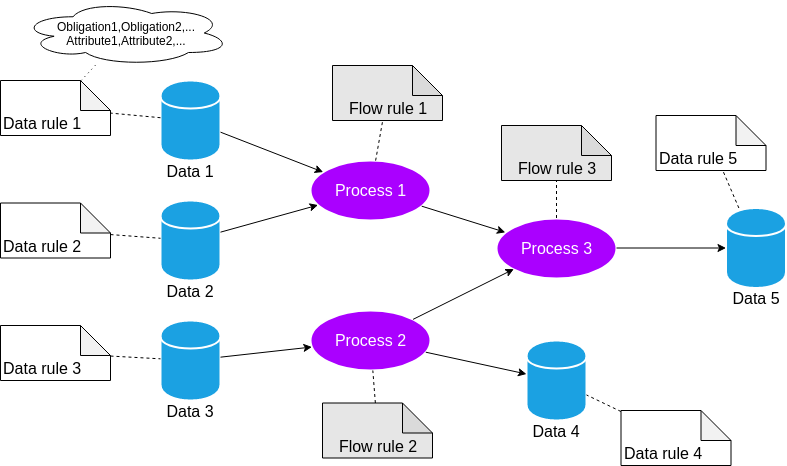}
    \caption{Annotation targets of \tterm{data rules} and \tterm{flow rules}}
    \label{fig:rule-target}
\end{figure}

\noindent Another language is also presented to model the flow behaviors, due to the necessity of referring to the governance rules. We call this the \tterm{flow rule} (language), and call the modelled governance rules \tterm{data rule}. As the name shows, the \tterm{data rule} is bound to the data, while the \tterm{flow rule} describes the process, as shown in Fig.~\ref{fig:rule-target}.

In the following part, we first introduce our differentiation of ``obligation and obligation definition''; the motivation and design of the \tterm{data rule} and \tterm{flow rule} languages are described below; the last subsection presents encoding examples.

\subsection{Obligation and obligation definition}
\label{sec:ob-ob-def}

An underlying concern across our work is the differentiation between ``obligation'' and ``obligation definition''. This is introduced here to avoid confusion.

An ``obligation'' is an action that an agent (often a human) should perform. It can be bound to the agent, a piece of data, etc. It is an \emph{action} that must be performed (otherwise it will be a violation).

On the other hand, an ``obligation definition'' specifies ``what obligation will be activated under what situation''. It is not an ``obligation'' itself, but describes how an obligation will be created.

Therefore, a key difference is that ``obligation definitions'' flow with data, while ``obligations'' may be ``global'' information. In fact, the current language only supports global\footnote{``Global'' in the \tkw{session} (see section \ref{sec:dr-session}).} obligations.

To keep it short, this paper generally doesn't distinguish between these two terminologies when no confusion can be made. When necessary, we use ``obligated action'' to described ``obligation''.

\subsection{Data rule}
\label{sec:data-rule}

The data rule describes the governance rules of a dataset. It should capture the different aspects that a governance rule may talk about.

A large aspect missed in most of the reviewed related work is \tterm{obligation}. Examples include ``\trule{the third column of the dataset contains IP addresses, and should not be leaked to the outside}'' and ``\trule{due to funding concerns, use of our data should be reported back to us}''.

As shown in the examples above, the ability to describe ``obligation'' is important to model governance rules. Having this mechanism, we can mimic other types of rules (e.g.\ privacy control).

In our work, the concept of \tterm{obligation} is borrowed from \cite{elrakaiby_formal_2012}: the action that the data processor (normally human) needs to do. Specifically, the \emph{post obligations} and \emph{privacy policies} are the main focus of the current work.

In the original model, the obligation in \cite{elrakaiby_formal_2012} can contain an \tterm{identifier}, a \tterm{subject} (or a \tterm{role}) who the obligation is related to, an \tterm{action} which should be executed (i.e.~the obligation), an \tterm{object} on which the obligation acts on, an \tterm{activation context} (condition) and a \tterm{violation context} (condition).

Our model extends their model and makes adjustments to provide more freedom in writing the obligations and remove the unnecessary parts. In our work, the \tterm{identifier} need not be supplied (it is automatically generated), because the activated obligations will be stored as a list; the \tterm{object} is a part of the specification of the obligation; the \tterm{violation context} is not implemented at this stage. The \tterm{subject} of the obligation is considered as the data processor (assumed to be the person who started the session), so its specification only defines who the obligation may apply to (like a filter) and it can be merged into the \tterm{activation context}\footnote{The working context requires a user identification at the start of session. User who initiated action, such as running a specific workflow, is then the subject. This is recorded in the provenance and can be easily used.}. Therefore, the aspects that the obligation definition will need to talk about are:

\begin{itemize}
    \item \tkw{obligated action} \quad the action that the data processor needs to do when this obligation is activated;
    \item \tkw{validity binding} \quad the data\footnote{The ``data'' here is actually \tkw{attribute} (as described in the latter parts).} dependency affecting the obligation definition;
    \item \tkw{activation condition} \quad the condition that triggers the obligation.
\end{itemize}

The \tkw{obligated action} in our model contains both the \tterm{action} and \tterm{object} in the previous model, with more freedom to add additional information; the \tkw{validity binding} is a new concept in our model, specifically designed for the data-flow oriented point of view; the \tkw{activation condition} is almost the same concept as the \tterm{activation context}, with the addition of \tterm{subject} as a condition. In addition, to have a better syntax, \tkw{attribute} is introduced mainly to facilitate \tkw{validity binding} when used in flow rules (see \ref{sec:flow-rule}). They are all explained below.

\subsubsection{Obligated action}
\label{sec:dr-action}

To facilitate reuse and interoperability, we propose the \tkw{obligated action} to be an instantiation\footnote{The correct word should be ``individual''. But we use ``instantiation'' and ``instance'' to mean they comply with the OWL axioms defined in the class and only exist / make sense in the context of our system.} of a \tterm{class} defined in an ontology. The arguments used to instantiate will also be specified in the data rule, using the \tkw{attribute} mechanism.

\subsubsection{Validity binding}
\label{sec:dr-binding}

The \tkw{validity binding} is the data (or the part of the data) that this obligation applies to, and defaults to the whole dataset. If present, it refers to an \tkw{attribute}, and the meaning is: ``this obligation definition is in force if and only if that specific attribute exists''.

\subsubsection{Activation condition}
\label{sec:dr-ac-cond}

This element specifies the condition on which the obligation will be activated. Activated obligations will be handled separately (stored into a list, in the current implementation), but the obligation definition remains. Ideally, users should check the list of activated obligations after processing and conduct the activated obligations.

\subsubsection{Attribute}
\label{sec:dr-attr}

\tkw{Attributes} are the extra information in the data rule, like \tterm{variables} in programming. They can be used (referenced) by the \tkw{obligated action} and the \tkw{validity binding}. An \tkw{attribute} contains \begin{enumerate*}[label=\alph*)]
    \item an \tterm{identifier} (or id for short);
    \item a \tterm{value struct}.
\end{enumerate*}
The identifier is an IRI (Internationalized Resource Identifier) and can be referenced (from \tkw{obligated action} and \tkw{validity binding}; the \tterm{value struct} is a struct (as in programming). In fact, there may be multiple \tkw{attributes} with the same \tterm{identifier}, so they will be merged into an ordered set of \tterm{value structs}, and be referenced with index.

\subsubsection{Session}
\label{sec:dr-session}

\tkw{Session} determines the checking scope of some \tkw{activation conditions}. For example, the \ttdrin|:OnImport| activation condition is \texttt{true} only when the rule appears in an \tterm{initial component} (see \ref{sec:impl}).

In the current implementation, \tkw{session} is handled very naively: one provenance graph is considered as one session. Normally, one execution of a workflow (in a workflow execution system supporting provenance) generates one provenance graph, so we can also say one workflow execution is one session. In practice, users run many workflows per sessions but sometimes re-attach a new session to a running workflow. This complexity is not relevant here.

\subsection{Flow rule}
\label{sec:flow-rule}

Flow rules describe how the data rules propagate through a process in a data-flow graph. Therefore, the flow rules of all processes describe how the data rules of the data-flow graph inputs propagate to the outputs.

The flow rule of a process should reflect the important aspects of the actual data processing. For example, which outputs are related to which inputs, what data transformation has been done, etc.

The flow rule consists of two parts:

\begin{itemize}
    \item \tterm{port mappings};
    \item \tterm{refinements}.
\end{itemize}

\noindent The \tterm{port mappings} are the general behavior of how the data rules flow from inputs to outputs. They come in the form of input-output mappings (such as $(inputN, outputM)$), and should be interpreted as ``every data rule from input N goes to output M (but may be refined)''. The default mapping is ``every data rule from every input port goes to every output port''.

The (data) rule at the output may emerge modified to reflect processing. The modifications act on the \tkw{attributes}, and can be either \emph{delete} or \emph{edit}. The reason why \emph{add} is not allowed at the moment is because \tkw{attributes} take effect only when referenced in obligations (so adding an dangling attribute makes no sense)\footnote{Generally, an obligation (definition) is atomic but carries modifiable \tkw{attributes}. If the processing adds a significant new requirement that is shown by including a new obligation on an output. This is beyond the design of modifying \tkw{attributes}.}.

The refinements need to specify the \tkw{attributes} they apply to. If the targeted \tkw{attribute} does not exist, this refinement simply passes without doing anything.

A \emph{delete} refinement can either act on all \tkw{attributes} with a specified id, or a particular \tkw{attribute} with a specified \tterm{value struct}. On the other hand, an \emph{edit} refinement can only act on an \tkw{attribute} with a specified id and a specified \tterm{value struct}, and it needs to specify a new \tterm{value struct} as the new value. Both \emph{delete} and \emph{edit} should specify which output port they are acting on and (optionally) which input port the attribute is from.

\begin{figure*}
    \centering
    \includegraphics[width=0.9\linewidth]{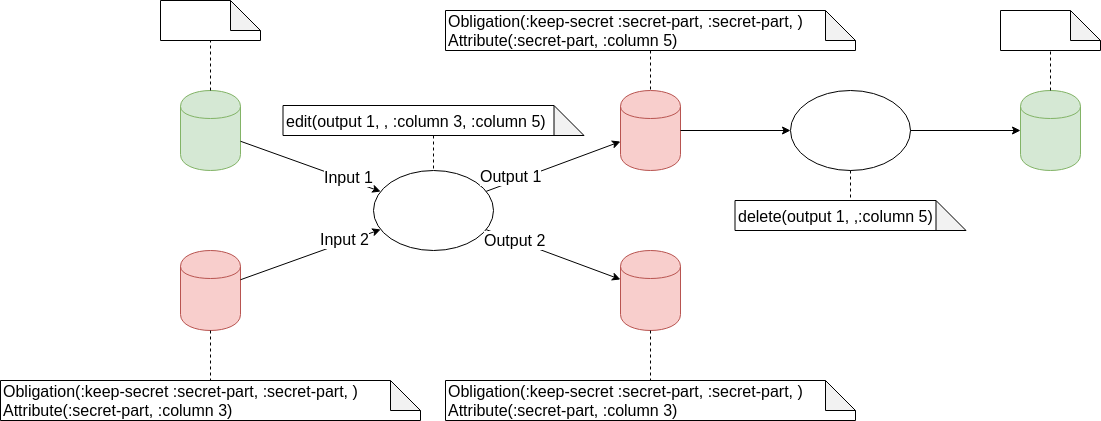}
    \caption{An example demonstrating the use of flow rules: here they show adaption of the data rule with the column number change. Ellipse for processes; cylinder for data; note-shape for rules; green for data without sensitivity; red for data with sensitivity; the flow rules here follow the default propagation (i.e.~everything in to everything out).}
    \label{fig:flow-rule-example}
\end{figure*}

With these mechanisms, it is possible to capture some useful information that would otherwise not be possible. For example, the 3\textsuperscript{rd} column of input 2 is placed at the 5\textsuperscript{th} column in output 1. Moreover, if a consecutive process removes column 5, we can now conclude the original 3\textsuperscript{rd} column is removed in the output of the second process. This is especially useful if the original 3\textsuperscript{rd} column contains sensitive information. Fig.~\ref{fig:flow-rule-example} demonstrates this example.

After the refinement, \tterm{merging} is conducted. The very basic intuition for \tterm{merging} is: it is possible that rules from different input ports go to the same output port and these rules may contain the same \tkw{obligated action} class or the same \tkw{attribute} id (for various meaningful reasons, such as they were originally from the same dataset). Therefore, the objective of \tterm{merging} is to remove redundant rules and match \tkw{attribute} references.

Merging is essentially the merge of sets (the obligation set and the attribute set) between two (or more) resources. But because the obligation set contains references to the attribute set, it may need to be slightly altered. It executes as follows (see Algorithm \ref{alg:merge} for a pseudo-code description):

\begin{enumerate}
    \item for all attributes with the same id, remove redundancy and redirect the references (in the obligations) to the corresponding remaining attributes;
    \item for all obligations, remove the obligations that are the same (i.e.~having the same action, validity binding and activation condition)\footnote{The \emph{equality} is in the ontology sense: the whole IRI (including the namespace) should match. A more extended solution should take ontological reasoning into account (especially the \texttt{owl:sameAs} axiom), because two different bodies may use different ontologies and parameters to express the same concept.};
    \item return the remaining attributes and obligations.
\end{enumerate}

\begin{algorithm}
    \caption{Merging algorithm (for each output port)
    \label{alg:merge}}
    \begin{algorithmic}[1]
    \REQUIRE{$rs$ is the list of rulesets (of data rules) to be merged}
    
    \LET{$o$}{empty ruleset}
    
    \FOR{$r$ in $rs$}
        \STATE add all in $r.obs$ to $o.obs$
        \FOR[attributes]{$attr$ in $r.attrs$} 
            \IF[both the same id and value]{$attr$ in $o.attrs$}
                \LET{$new\_attr\_ref$}{reference to $attr$ in $o.attrs$}
                \FORALL{$ob$ in $o.obs$ such that $ob$ references $attr$}
                    \STATE update reference of $attr$ to $new\_attr\_ref$
                    \IF{$ob$ has duplicates}
                        \STATE remove $ob$ from $o.obs$
                    \ENDIF
                \ENDFOR
            \ELSE
                \STATE add $attr$ to $o.attrs$
            \ENDIF
        \ENDFOR
    \ENDFOR
    \RETURN $o$
    
    \end{algorithmic}
\end{algorithm}

\subsection{Encoding examples}

In this section, we demonstrate the encoding examples of some governance rules as obligations, including the two example rules given at the beginning of \ref{sec:data-rule}.

For simplicity, we omit the ontology prefixes (namespaces). In fact, they are not used for special purposes (except for identifier) for this paper\footnote{In the experiment, we developed a very primitive ontology. The ontology to use in a production system is still under development to provide better interoperability. The ontology to describe internal structures of data is external to our research.}.

\subsubsection{Keep third column secret}

For the rule ``\trule{the third column of the dataset contains IP addresses, and should not be leaked to the outside}'', the key information is that the specific column should be kept secret, so this can be encoded as:

\begin{lstlisting}[style=datarule]
Obligation(:secret :col3, :col3, )
Attribute(:col3, :column 3)
\end{lstlisting}

Here, \ttdrin|:secret| and \ttdrin|:column| are two ontology classes. We omit the prefixes (namespaces) but we keep the \ttdrin|:| to indicate this should be an ontological reference. The \ttdrin|:col3| is the id of an individual as a result of the \ttdrin|Attribute()| statement\footnote{From the computational point of view, it is automatically created in the system.}.

The \ttdrin|Attribute()| statement defines an attribute, with id \ttdrin|:col3|, and value \ttdrin|:column 3|. This id is the individual IRI, so it can be referenced in the \ttdrin|Obligation()| statement\footnote{In fact, this attribute id doesn't have to be in the ontology world because it is not persistent. However, to make the syntax explicitly show this is not a \emph{value}, we keep the \ttdrin|:| here and describe it as an individual for simplicity.}.

The \ttdrin|Obligation()| statement contains three elements, separated by comma. The first element is the \tkw{obligated action} and the form \ttdrin|:secret :col3| means an instance of the ontological class \ttdrin|:secret|, and the argument is the (dereferenced) value of (the attribute) \ttdrin|:col3| (i.e.~\ttdrin|:column 3| initially). The second element is the \tkw{validity binding}, bound to the (same) \ttdrin|:col3| \tkw{attribute}. The third element is the \tkw{activation condition}, but is empty for this obligation.

An empty \tkw{activation condition} means this obligation will never be activated. As a consequence, we use this as a special case meaning the related information should be checked after processing (e.g.~check all output data to see if any of them have the \ttdrin|:secret| obligation). A better syntax may be used as a future work.

\subsubsection{Report data use}

The stem of the rule ``\trule{due to funding concerns, use of the data should be reported back to us}'' is actually ``\trule{use of the data should be reported back to us}'', so this can be encoded as:

\begin{lstlisting}[style=datarule]
Obligation(:report :source, :source, :OnImport)
Attribute(:source, "Fictional source")
\end{lstlisting}

The \ttdrin|:OnImport| is the \tkw{activation condition} of this obligation, meaning this obligation will be activated when the data is first introduced to the execution session.

Specifically, our rule language enables the data provider to be explicit about what they mean by ``use''. The example above defines ``use'' as ``the first time as input to the workflow''. An alternative definition could be ``every time it is sent as an input to a component'':

\begin{lstlisting}[style=datarule]
Obligation(:report :source, :source, :OnAsInput)
Attribute(:source, "Fictional source")
\end{lstlisting}

\subsubsection{Acknowledge when publish}

A common rule used in research is ``\trule{if you use our dataset and make a publication based on it, you should acknowledge this in your publication using a proper way such as `XX'}''. This could be encoded as:

\begin{lstlisting}[style=datarule]
Obligation(:acknowledge :form, , :OnPublish)
Attribute(:form, "XX")
\end{lstlisting}

The \ttdrin|:OnPublish| condition matches a component which is identified as making a publication. This component is actually a ``virtual'' component, meaning it is not a component in the provenance (of the executed workflow), but an additional element added to the provenance at runtime to assist the reasoning.

%% file: sec_system.tex
\section{System}
\label{sec:arch}

\begin{figure}
    \centering
    \includegraphics[width=\linewidth]{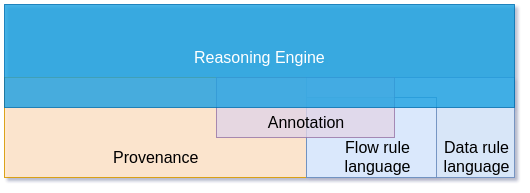}
    \caption{System architecture}
    \label{fig:system-arch}
\end{figure}

Facing the research questions and taking lessons from existing research, our system provides a mechanism to model more than the access control aspect of the data governance rules in a federated context, using the rule language described in section \ref{sec:lang}. In this section, we describe the system architecture (Fig. \ref{fig:system-arch}) and the design concerns.

\subsection{Architecture}

Since the system intends to provide a vendor- and technology-agnostic method to deal with rules, the foundation must have the same neutrality. The foundation contains two parts:
\begin{enumerate*}[label=\alph*)]
    \item data-flow representation; and
    \item rule encoding language.
\end{enumerate*}
The rule languages are already described in section \ref{sec:lang}. Hence, this section mainly describes the data-flow representation.

The data-flow representation is the main input to our system (plus the encoded rules). Therefore, we choose \emph{provenance} W3C PROV \cite{groth_prov-overview._2013} to be the representation, taking the advantage that PROV-O builds on OWL \cite{mcguinness_owl_2004} which allows the interoperability between different institutions by technologies like ontology matching. In fact, as mentioned above, our rule language also uses ontology for interoperability.

Specifically, we use S-Prov \cite{spinuso_active_2018}, an ontology extending W3C PROV-O to record data-streaming workflow executions, as the provenance specification. The reason we choose this ontology is because
\begin{enumerate*}[label=\alph*)]
    \item it provides necessary information to identify workflow components;
    \item our architecture is data-centric which aligns with the data-streaming perspective of S-Prov.
\end{enumerate*}
However, the data-streaming feature is not used in our system for the moment, so in principle our system can also take other ontologies (e.g. OPMW\footnote{OPMW-PROV: The Open Provenance Model for Workflows http://www.opmw.org/index.html}) as the input, either by directly having a separated parser or using ontology matching from the other representation to S-Prov.

Thus, our system uses the provenance to reconstruct the data flow and identify relevant information (e.g.~the agent who initiated the execution), as the information necessary for reasoning (e.g. in the \tkw{activation condition} of data rules).

Because our usage context is to help researchers comply with the rules, there is no need to ``enforce'' the rules in the system. Instead, a method such as a prompt is considered enough for the user (i.e.~researcher) either as a reminder or a checklist. For example, a user can ask what obligations he has pending.

Therefore, the reasoning engine takes both the information from provenance and the rules as input to reason about:

\begin{enumerate}
    \item the governance rules of the output data at each stage; and
    \item the activated obligations.
\end{enumerate}

These two missions are essentially the reason why we design these two sets of rules -- they can be expressed as:

\begin{enumerate}
    \item conduct the flow rules;
    \item check the activation conditions of data rules (and instantiate the activated obligations).
\end{enumerate}

\subsection{Implementation details}
\label{sec:impl}

Because the rule languages are not yet expressed in logic, we are unable to use formal reasoners (theorem provers, for example). Instead, we developed a computational reasoning system as proof-of-concept to demonstrate the feasibility of this reasoning since the semantics of the rule languages are clear.

The system takes the provenance as the input and then extracts the data flow to construct a directed graph -- vertices are the processing steps (components) and the edges are the data transmission between components. SPARQL is used for this procedure, and the extra information like the \texttt{component function} is kept in the resulting graph. The flow rules are also inserted into the resulting graph.

After obtaining the graph, the reasoning can be conducted for every node with 0 in-degree and then repeatedly for the rest of the graph with 0 in-degree. This is essentially the same as doing a topological sorting based on the in-degrees and then performing reasoning from the beginning to the end.

For every component, the reasoning does the following:

\begin{enumerate}
    \item Receive the data rules for every input port;
    \item Identify the current context of execution and check the activation conditions of every data rule;
        If any activation condition is met, instantiate and store the activated obligation;
    \item Propagate the data rules from the inputs to outputs in accordance with the flow rules.
\end{enumerate}

In our implementation, the data rules are stored as an augmentation to the graph extracted from provenance, directly attached to the output ports. The reason why they are not attached to data is because S-Prov schema aims at the data-streaming style processing, so data are split into small chunks and therefore data is produced and transmitted from an output to an input incrementally. We assume that the governing rules are the same for every data unit in the stream.

Specifically, the data rules for the \emph{initial components} (i.e.~components that do not take any inputs) are \tterm{imported}. We use the so-called \tterm{recognizer} to identify what rules shall be \tterm{imported} by these components, and assign a fake input with these rules. In a real production system, this should be replaced by a more sophisticated way, but it is enough for the proof-of-concept.

\subsection{Initial results}

We conducted a few experiments to verify the system's correctness and the model's feasibility.

We assume the data flow graphs are always fully connected Directed Acyclic Graphs (DAGs) with dangling input and output ports / edges (which are directed connected to input and output data). Therefore, the components are the vertices of the graph, the data-flow connections are the edges. 
The potential topology of DAGs is defined by its vertices and edges, and the type of vertices (in terms of edges) is finite, so we only need to ensure the correctness of our system for every one of these vertices (components):% (In fact, \ref{itm:vt:11}-\ref{itm:vt:n1} are all special cases of \ref{itm:vt:nn})

\begin{enumerate}
    % \item zero-input;
    \item \label{itm:vt:11} one-input-one-output (1-1), to test the propagation;
    \item \label{itm:vt:1n} one-input-multi-output (1-n), to test spreading;
    \item \label{itm:vt:n1} multi-input-one-output (n-1), to test the aggregation;
    \item \label{itm:vt:nn} multi-input-multi-output (n-m), to test redispatching.
\end{enumerate}

\noindent In principle, in a DAG, a vertex can have zero inputs or zero outputs. Let aside the ones with dangling edges, semantically, a component with zero inputs is a producer: it does not rely on external resource, but will produce data (e.g.~a prime number generator); a component with zero outputs makes no sense in a data flow (we argue a ``storage'' component also produces outputs) so we do not consider it at all. For experimental purposes, we use producers with one output port as the initial vertices, and connect the reset of the graph to them.

\noindent The synthetic workflows are written in dispel4py \cite{filguiera_dispel4py:_2014}, a data-streaming workflow execution system. We executed these workflows with provenance generation (coupled with S-ProvFlow \cite{spinuso_active_2018}), and then conducted reasoning. The reasoning intends to check that the propagation works correctly, so placeholders (simple strings) are used instead of meaningful data rules.

\begin{table}[]
    \centering	
    \caption{Pattern coverage of each synthetic workflow}
    \label{tbl:workflow-patterns}
    \begin{tabular}{llllll}
		Graph & 1-1 & 1-n & n-1 & n-m \\
		\hline
		single line streaming & \checkmark & & & \\
		spreading             & \checkmark & \checkmark & & \\
		collecting            & \checkmark & & \checkmark & \\
		redispatching         & \checkmark & & \checkmark & \checkmark
	\end{tabular}
\end{table}

\begin{figure*}
    \centering
    \includegraphics[width=\linewidth]{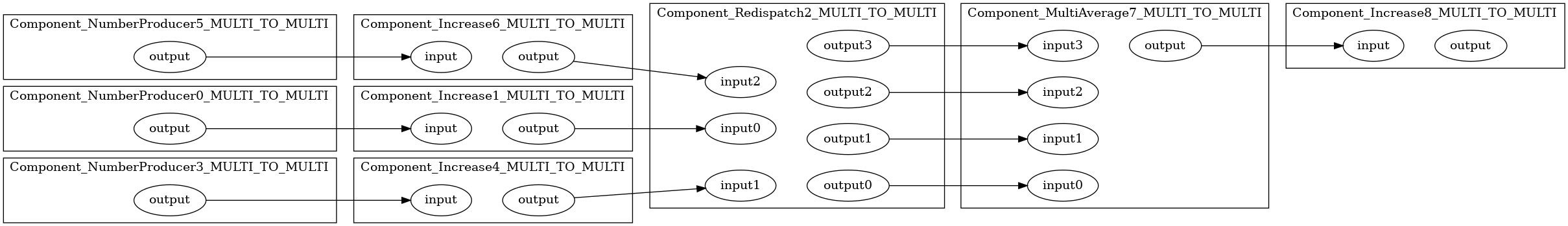}
    \caption{Topology of the ``redispatching'' synthetic workflow. \quad Squares are components; ovals are ports; arrows are connections.}
    \label{fig:workflow-multi-multi}
\end{figure*}

\begin{figure*}
    \centering
    \subfloat[left-half]{%
        \includegraphics[width=\linewidth]{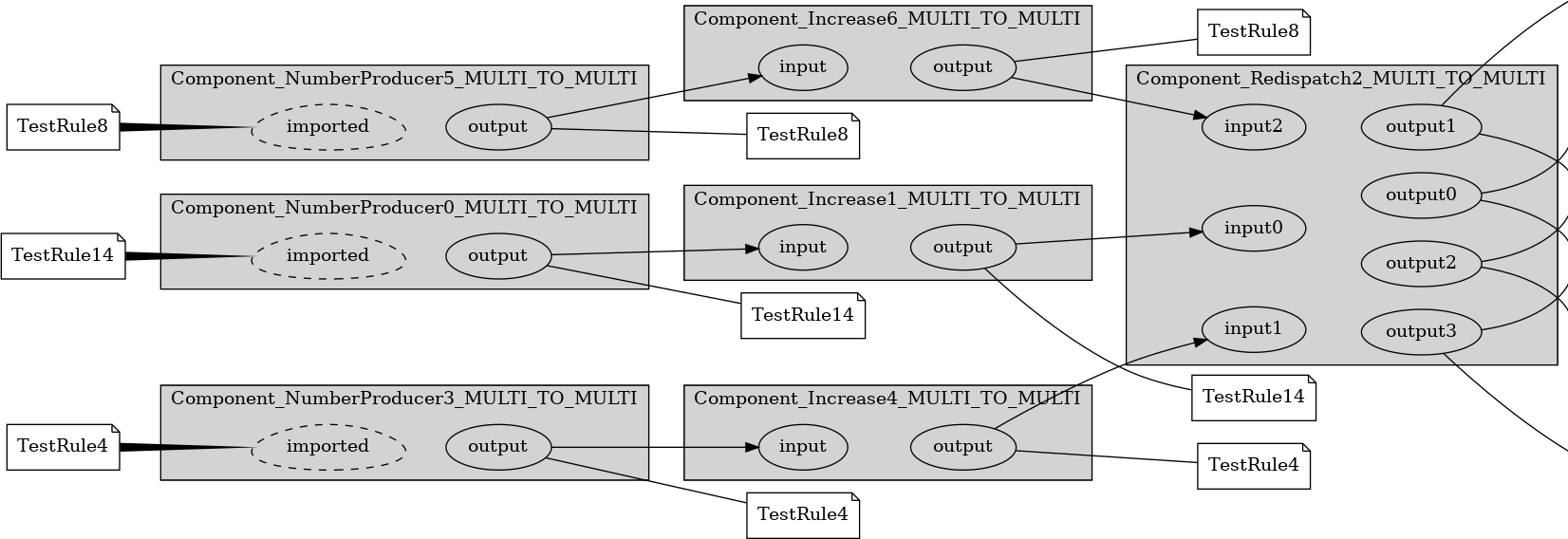}%
        }
        % \caption{left-half}
    \\
    \subfloat[right-half]{%
    % \begin{subfigure}{\linewidth}
        \includegraphics[width=\linewidth]{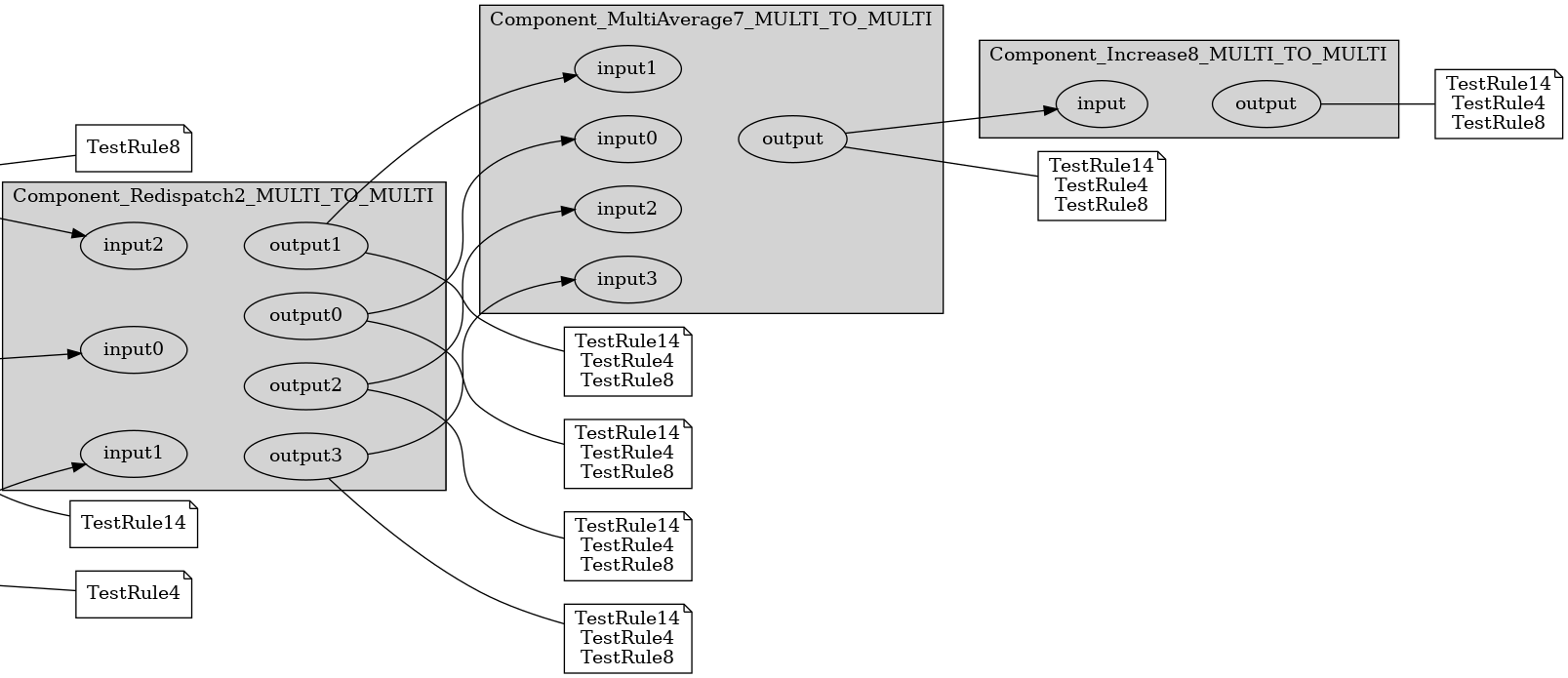}%
        }
    %     \caption{right-half}
    % \end{subfigure}
    % \includegraphics[width=\linewidth]{multi-to-multi-with-rules-v2-right.png}
    \caption{Reasoning result based on the provenance from executing the ``redispatching'' workflow (Fig.~\ref{fig:workflow-multi-multi}). \quad Due to its width, the figure is split into two halves (sharing \texttt{Component\_Redispatch2\_MULTI\_TO\_MULTI}): subfigure (a) is the left-half; (b) is the right-half. In addition to Fig.~\ref{fig:workflow-multi-multi}: note-shapes are rules; lines (without arrows) connect the ports and the corresponding rules; black triangle arrows indicate \tterm{imported rules}; egg-shapes indicate virtual port; components are greyed.}
    \label{fig:workflow-multi-multi-result}
\end{figure*}

The vertex patterns checked by each synthetic workflow are shown in Table \ref{tbl:workflow-patterns}. As a particular example, Fig.~\ref{fig:workflow-multi-multi} demonstrates the structure of the original synthetic workflow of \emph{redispatching}, and Fig.~\ref{fig:workflow-multi-multi-result} shows the reasoning result.

The success of these experiments demonstrates the correctness of our system and model. It also demonstrates the feasibility of trying to handle data governance rules in any DAGs.

%% file: sec_contribution.tex
\section{Contribution and future work}
\label{sec:contrib}

We describe the contribution of this paper and the future work in this section.

\subsection{Contribution}

The main contribution of this paper lies in five aspects:

\begin{itemize}
    \item clarified the questions concerning automating data-governance rules handling and identified the research targets in details in the \emph{federated context};
    \item demonstrated the use of provenance data as a source for rules;
    \item extended and refined the obligation model in \cite{elrakaiby_formal_2012} to better represent data-flow oriented data governance rule (policy) specification and the \emph{federated context};
    \item developed a method to describe the flow behavior of the rules for each process in a \emph{multi-input-multi-output} data flow graph;
    \item pioneered a systematic method to help researchers comply with data governance rules in a federated research community.
\end{itemize}

\subsection{Future work}

Since we are addressing a field which has received little attention, there is a lot of research needed.
This lies in three directions:

\begin{enumerate}
    \item Ground the rule languages into logic;
    \item Extend the language to capture more aspects;
    \item Develop supporting technologies.
\end{enumerate}

The following part describes these directions in more detail with our emphases. It's worth noting that these three directions interact. 
Specifically, explicit encoding of \emph{requirements} (mentioned in section \ref{sec:lang}) is related both to directions 2 and 3 (explained below): making the semantics explicit and providing a specific syntax in direction 2; providing the mechanism to check in direction 3.

\subsubsection{Logic grounding}
This direction will be our main focus in the next step, because 
\begin{enumerate*}[label=\alph*)]
    \item we can become more certain that there is no internal problem with the rule language;
    \item existing formal reasoners can be used, taking the advantage of their proven correctness;
    \item the rules from different sources can be compared by checking their logical conflicts, enabling more use cases.
\end{enumerate*}

The data rules and flow rules shall be grounded into two logic sets, because of their different properties. For any formalisation, the data rules should always be harmonious with OWL because ontology (and therefore OWL) is critical to interoperability. Similarly, because our framework builds on top of semantic technology, the modelling mechanism provided by OWL axioms is also worth exploring to encode and validate flow rules.

\subsubsection{Language extension}
This direction is also important and will be our secondary target. More fine-granulated semantic meanings can be encoded with an extended language. This should better be conducted with direction 1.

One item of future work may be to extend the \tkw{activation condition} to accept more triggers; another potential item is to add more language semantics, such as capturing and denoting the \tterm{violation context}, or making \tterm{session} customisable. In the meantime, ODRL may be a candidate to encode some elements of data rules (e.g.\ \tkw{obligated action}).

\subsubsection{Supporting technologies}
This direction concerns the collaboration and system implementation scenario, and generally would be conducted with other researchers from application contexts.
For example, our collaborator in KNMI will bring useful use cases for our framework supporting earth-system consortia.

An explicit direction may be to define a standard protocol to retrieve rules with data; real-time processing (instead of the current retrospective analysis) should also be investigated.
This would enable alerts to be sent to users when new obligations emerge and eventually lead to warnings preventing serious rule infringement, exploiting active provenance \cite{SpinusoBC2DC2019}.

%% file: sec_conclusion.tex
\section{Conclusion}
\label{sec:concl}

In this paper, we have identified a requirement for improving rule management to help users and providers work well together in a federated context. We proposed a method to help by delivering appropriate encoding and automation. This includes a formal model to encode the data governance rules, and a companion model to describe how the governance rules will flow and change during processing. We then presented some example encodings of governance rules using our model, and presented our proof-of-concept system as well as its initial results to demonstrate feasibility. Finally, we highlighted our contribution as a conceptual and practical framework.

We believe our work points to an important research direction, and provides a good first step towards solving that. More work will follow, as described in the future work section.